\documentclass[10pt,letterpaper]{article}
\usepackage[top=0.85in,left=2.75in,footskip=0.75in]{geometry}

\usepackage{amsmath,amssymb}

\usepackage{changepage}

\usepackage{textcomp,marvosym}

\usepackage{cite}

\usepackage{nameref,hyperref}

\usepackage[nopatch=eqnum]{microtype}
\DisableLigatures[f]{encoding = *, family = * }

\usepackage[table]{xcolor}

\usepackage{array}

\newcolumntype{+}{!{\vrule width 2pt}}

\newlength\savedwidth

\raggedright
\usepackage[aboveskip=1pt,labelfont=bf,labelsep=period,justification=raggedright,singlelinecheck=off]{caption}

\makeatletter
\renewcommand{\@biblabel}[1]{\quad#1.}
\makeatother

\usepackage{lastpage,fancyhdr,graphicx}
\usepackage{epstopdf}
\fancyheadoffset[L]{2.25in}
\fancyfootoffset[L]{2.25in}
\begin{document}
\vspace*{0.2in}

\begin{flushleft}
{\Large
\textbf\newline{Open Source Infrastructure for Automatic Cell Segmentation} 
}
\newline
\\
Aaron Rock Menezes\textsuperscript{1,2,*},
Bharath Ramsundar\textsuperscript{2,**},
\\
\bigskip
\textbf{1} Department of Biological Sciences, BITS Pilani, K.K. Birla Goa Campus
\\
\textbf{2} Deep Forest Sciences
\\
\bigskip

* aaron.r.menezes@gmail.com

** bharath@deepforestsci.com

\end{flushleft}
\section*{Abstract}
Automated cell segmentation is crucial for various biological and medical applications, facilitating tasks like cell counting, morphology analysis, and drug discovery. However, manual segmentation is time-consuming and prone to subjectivity, necessitating robust automated methods. This paper presents open-source infrastructure, utilizing the UNet model, a deep-learning architecture noted for its effectiveness in image segmentation tasks. This implementation is integrated into the open-source DeepChem package, enhancing accessibility and usability for researchers and practitioners. The resulting tool offers a convenient and user-friendly interface, reducing the barrier to entry for cell segmentation while maintaining high accuracy. Additionally, we benchmark this model against various datasets, demonstrating its robustness and versatility across different imaging conditions and cell types. 


\section*{Introduction}
In biological and medical research, achieving precise cell segmentation is essential for tasks like cell counting, morphology analysis, and drug discovery \cite{bengtsson2004robust}. However, manual segmentation methods are laborious, subjective, and prone to errors, especially when working with low-contrast microscopy techniques like bright field microscopy or with low-resolution imagery. These challenges prompt the need for robust automated solutions for cell segmentation. In recent years, the UNet model has emerged as a promising tool for image segmentation tasks, leveraging deep learning techniques to learn effective segmentations \cite{ronneberger2015unet}.  Implementing and training such a model can be daunting for users who aren't familiar with programming. Our work simplifies this process with a modular implementation that shields users from technical complexities, whilst enabling easy and convenient application of advanced image segmentation techniques. Our aim is to improve ease-of-use of automated cell segmentation, in order to help advance research across various scientific domains. Through evaluations on open-source microscopy datasets, we compare our UNet model on various open-source microscopy datasets and present a comparison with the current state-of-the-art results. 

DeepChem,\cite{Ramsundar-et-al-2019}, is an open-source Python library aimed at scientific machine learning and deep learning, focusing on molecular and quantum datasets.  DeepChem's structure empowers the tackling of complex scientific challenges in areas such as drug discovery, bioinformatics, and physics. The organized framework of DeepChem has enabled it to be used for applications from molecular machine learning assessments using the MoleculeNet benchmark suite\cite{wu2018moleculenet} to protein-ligand interaction modeling \cite{gomes2017atomic}, and generative modeling of molecules \cite{frey2022fastflows}, among others. DeepChem currently has limited support for image-based data. This work addresses this shortcoming by improving DeepChem's image-handling functionality and leverages this improved functionality to build effective cell segmentation tools.

In particular, this work integrates the UNet model in addition to tutorials for automated cell counting and segmentation to enable users to analyze image data, e.g. microscopy or biomedical imagery. In addition to the integration of the UNet model, we have also improved support for image-based datasets by improving the implementation of the \textit{ImageLoader}, \textit{ImageDataset} and \textit{ImageTransformer} classes in DeepChem to facilitate the loading and pre-processing of image data.

\section*{Implementation}
We have built a full pipeline for analyzing microscopy images using DeepChem. This pipeline consists of the following parts: loading microscopy datasets using DeepChem's \textit{ImageLoader} class, creating an \textit{ImageDataset} and pre-processing the data, and then using the UNet model to train a model on the dataset.

\subsection*{Microscopy Datasets}
There are many open-source microscopy datasets such as LIVECell\cite{edlund2021livecell}, the Broad Bioimage Benchmark Collection\cite{ljosa2012annotated}, the ISBI Cell Tracking Challenge\cite{mavska2023cell}. To evaluate the generalizability of our model, we utilize a diverse dataset collection encompassing various microscopy techniques and biological subjects. This includes the publicly available BBBC003v1 and BBBC039v1 datasets from the Broad Bioimage Benchmark Collection\cite{ljosa2012annotated}, featuring DIC microscopy of Mouse embryos and fluorescence microscopy of the U2OS cell line of human osteosarcoma cells. We further leverage the open source datasets from the ISBI Cell Tracking Challenge\cite{mavska2023cell}, incorporating a wider range of microscopy methods (phase contrast, fluorescence, and differential interference contrast microscopy) and cell types (mouse stem cells, HeLa cells, pancreatic stem cells, etc.). This selection allows us to test our model's performance on distinct cell morphologies and imaging conditions, fostering a comprehensive assessment of its robustness.

\begin{table}[!ht]
\begin{adjustwidth}{-2.25in}{0in} 
  \centering
\caption{{\bf Summary of the datasets used in our benchmark experiments, with the types of cells imaged and microscopy techniques used to capture the images.}}
\vspace{8px}
  \begin{tabular}{|l|l|l|} 
  \hline
    \textbf{Dataset} & \textbf{Cell Type} & \textbf{Microscopy Technique}\\ \hline
    BBBC003 & Mouse embryos & Differential Interference Contrast \\ \hline
    BBBC039 & Huma osteosarcoma U2OS cells& Fluorescence \\ \hline
    DIC-C2DH-HeLa & HeLa cells & Differential Interference Contrast \\ \hline
    Fluo-C2DL-MSC & Rat mesenchymal stem cells& Fluorescence \\ \hline
    Fluo-N2DH-GOWT1 &  GFP-GOWT1 Mouse stem cells& Fluorescence \\ \hline
    Fluo-N2DH-HeLa & HeLa cells & Fluorescence \\ \hline
    Fluo-N2DH-SIM &  Human Leukemia HL60 cells& Fluorescence \\ \hline
    PhC-C2DH-U373 &  Glioblastoma-astrocytoma U373 cells& Phase Contrast \\ \hline
    PhC-C2DL-PSC & Pancreatic stem cells& Phase Contrast \\ \hline
  \end{tabular}
  \label{table1}
\end{adjustwidth}
\end{table}

\subsection*{Image Loaders, Datasets, and Pre-Processing}
Image Loaders and Image Datasets are an integral part of the DeepChem package and allow scientists to easily access image datasets to visualize and process their data. The \textit{ImageLoader} class can be used to load data from folders and create \textit{ImageDatasets}, which can be used to access, analyze, and pre-process image datasets as well as for training and testing models. We improved the ImageDatasets to be able to work with images as both inputs and labels, a functionality that was missing prior to this work. This enables practitioners and researchers to conduct a wide array of computer vision based experiments.

DeepChem also has several \textit{Transformers} which allow users to process data in various ways, such as using the \textit{NormalizationTransformer} for normalization of data or the \textit{ImageTransformer} for resizing of images.

\subsection*{UNet Model}
The UNet model architecture is structured as a symmetric encoder-decoder convolutional neural network. At its core, it comprises a contracting path, where each layer progressively downsamples the input image's spatial dimensions while increasing the number of feature channels. This contracting path consists of convolution layers followed by max-pooling operations, enabling the extraction of high-level features.

\begin{figure}[!ht]
    \centering
    \includegraphics[width=1\linewidth]{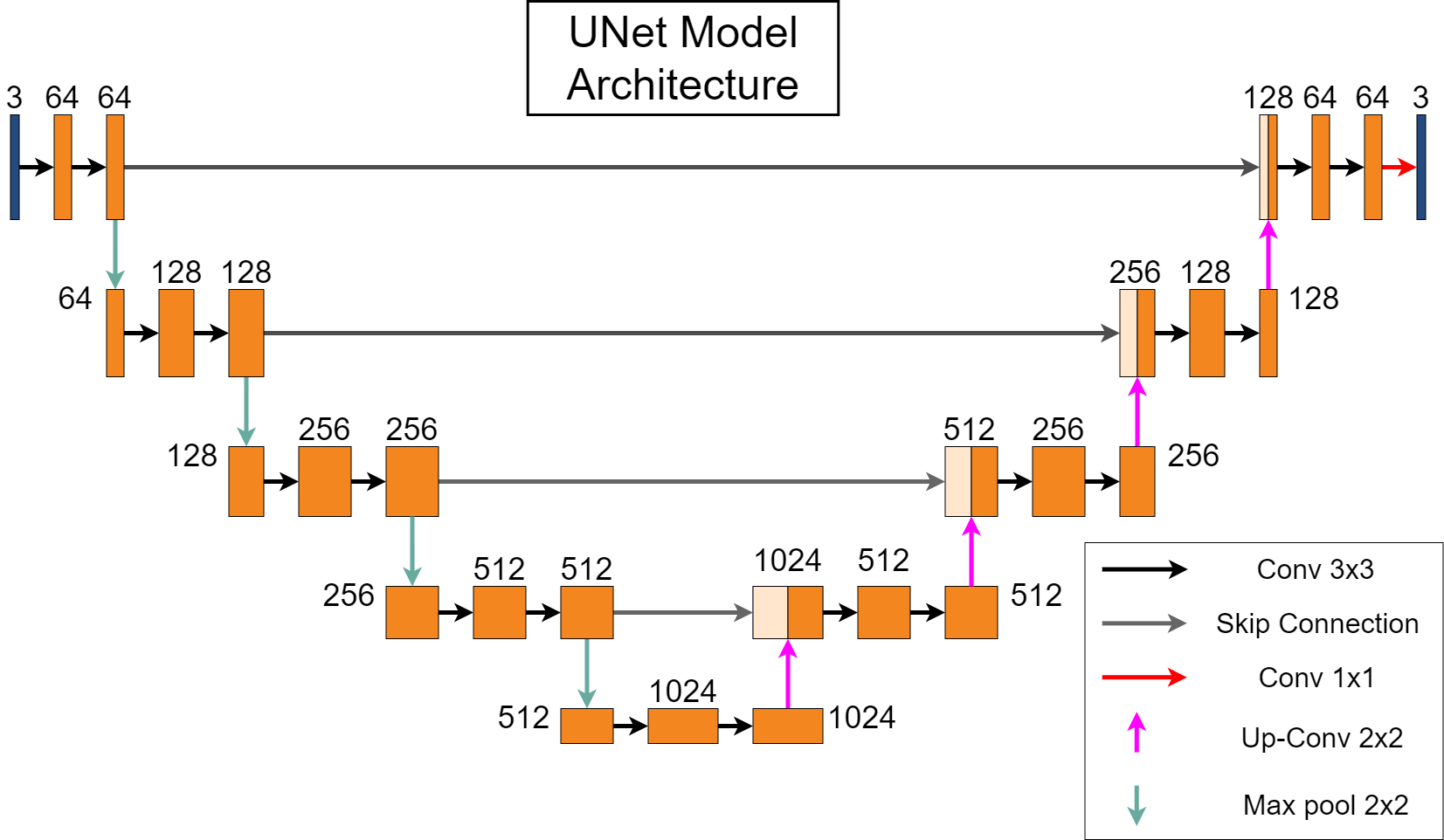}
    \caption{Block diagram of the UNet model architecture. The numbers show the number of channels in the image, we can see that our input and output are both 3-channel images. }
    \label{figure1}
\end{figure}

To facilitate information flow between corresponding encoder and decoder layers, skip connections are employed, which concatenate feature maps from the contracting path with those from the expansive path, aiding in the preservation of spatial information and enabling precise segmentation\cite{long2015fully, drozdzal2016importance}. Our UNet model, unlike the original implementation, pads all images to ensure spatial dimensions are consistent across all layers of the network to prevent any loss of information at the image borders. Our implementation allows users to select the number of input and output channels in addition to the optimizer, loss function, and the learning rate they wish to use.

Incorporating the UNet model into DeepChem involved utilizing PyTorch\cite{paszke2019pytorch} as the backend framework. We integrated the UNet model into DeepChem, enabling researchers to access and utilize this tool for automated cell segmentation and other related tasks.

\subsection*{Cell Segmentation Pipeline}
Loading and pre-processing of data, training of the model and evaluation and inference of the model follows a simple pipeline, which can be achieved in a few lines of Python code. We describe the pipeline in Figure 1, which involves downloading and loading the data as an \textit{ImageDataset} using an \textit{ImageLoader}, preprocessing using the various \textit{Transformers} followed by training and evaluating the UNet. 

\begin{figure}[!ht]
    \centering    \includegraphics[width=1\linewidth]{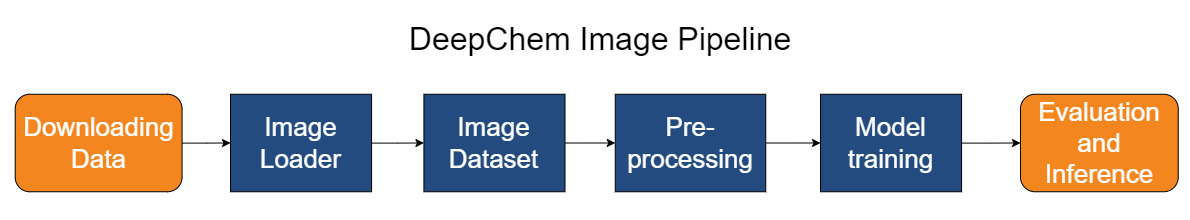}
    \caption{Overview of the cell segmentation pipeline using DeepChem. The pipeline includes data loading, pre-processing, model training, evaluation, and inference. }
    \label{figure2}
\end{figure}

Users can download and process their data, and train models on it using just a few lines of code without having to refer to various packages or guides.   

\begin{figure}[!ht]
    \centering
    \includegraphics[width=1\linewidth]{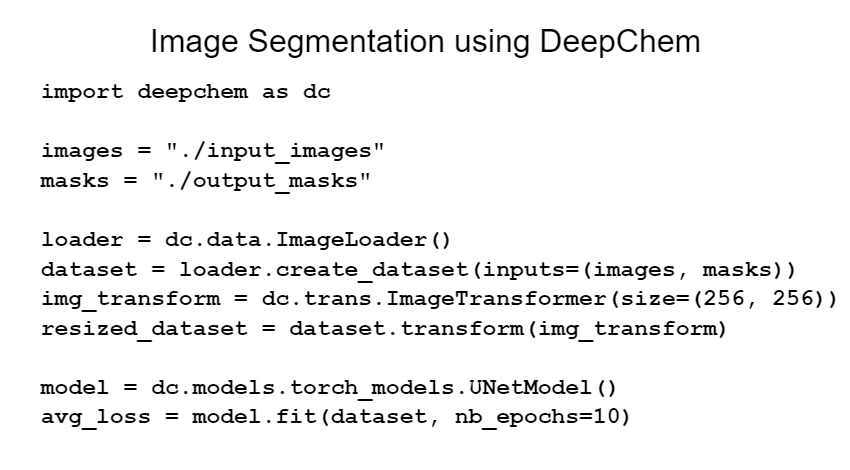}
    \caption{This is the DeepChem implementation of the Image Segmentation Pipeline as seen in Fig. 2.}
    \label{figure3}
\end{figure}

\begin{table}[h]
    \centering
\caption{Comparison of the lines of code written to implement our pipeline using DeepChem and using only PyTorch\cite{buda2019association}. }
\vspace{8px}
    \begin{tabular}{|l|l|l|} \hline 
          &DeepChem & PyTorch \\ \hline 
          Lines of Code&9& 250-300 \\ \hline
    \end{tabular}
    \label{table2}
\end{table}

\section*{Results}

\subsection*{Experimental Setup}
We test the UNet model on various open-source datasets. All code was run using Google Colab Pro, on 1 NVIDIA A100 GPU. We report results averaged across 5 runs of the datasets mentioned above. For preprocessing the data, we normalize the pixel intensities of all the input images and ensure that the segmentation masks were binary in nature. All images are scaled down to a height of 256 and the width was scaled in proportion to the image's original aspect ratio to the nearest multiple of 16. We use a batch size of 2 and train the models for a total of 100 epochs. All models are trained using Binary Cross Entropy Loss and the Adam optimizer with a learning rate of \(10^{-4}\). The models are evaluated using Intersection over Union (IoU), the F1 Score and the Area under ROC (AuROC) as metrics.

\subsection*{Experimental Results}

Models trained on the BBBC003 and BBBC0039 datasets from the Broad Bioimage Benchmark Collection show good performance using an 80-20 train-test split. These datasets were fairly small (16-200 images) compared to the Cell Tracking Challenge's datasets.  

\begin{table}[!ht]
    \centering
\caption{Evaluation performance of the DeepChem UNet Model on BBBC003 and BBBC039.}
\vspace{8px}
    \begin{tabular}{|l|l|l|l|l|l|} 
    \hline 
         Dataset&  Precision & Recall&  F1 Score& AuROC& mIoU\\ \hline 
         BBBC003&  0.7624& 0.8263&  0.7930& 0.9888& 0.6571\\ \hline 
         BBBC039&  0.9086& 0.9902&  0.9477& 0.9989& 0.9006\\\hline
    \end{tabular}
    \label{table3}
\end{table}

We also trained the model on a subset of the Cell Tracking Challenge's datasets. Each dataset contains 2 sets of images of cells taken over a duration of time at regular intervals. As these datasets contain images of cells over a fixed duration of time, we treated each frame as an independent image and randomly split the data for each sequence. We trained and tested the model on the 2 different sequences for each dataset due to the unavailability of the test set's labels. 

\begin{table}[!ht]
\begin{adjustwidth}{-2.25in}{0in}
    \centering
\caption{Evaluation performance of the DeepChem UNet Model on datasets from the Cell Tracking
Challenge 2020. Each Cell Tracking dataset has 2 captured sequences of images. We trained models
on sequence 1 and tested on sequence 2 for each dataset, as the test set is not public. SOTA numbers
are from the test set and are not directly comparable with our benchmarks but serve as a useful
comparison point.}
\vspace{8px}
    \begin{tabular}{|l|l|l|l|l|l|l|l|}
    \hline 
        Dataset & Precision & Recall & F1 Score & AuROC & mIoU & SOTA mIoU\cite{cell_segmentation_benchmark}\\ \hline 
        DIC-C2DH-HeLa& 0.8977& 0.7841&  0.8371& 0.9327&  0.7198& 0.877\\ \hline 
        Fluo-C2DL-MSC& 0.8375& 0.8059&  0.8214& 0.9698&  0.6969&  0.687\\ \hline 
        Fluo-N2DH-GOWT1& 0.9742& 0.8970&  0.9340& 0.9943&  0.8762&  0.938\\ \hline 
        Fluo-N2DH-HeLa& 0.9476& 0.9451&  0.9463& 0.9932&  0.8982&  0.923\\ \hline 
        Fluo-N2DH-SIM& 0.8709& 0.5915&  0.7045& 0.9395&  0.5438&  0.832\\ \hline 
        PhC-C2DH-U373& 0.8854& 0.9259&  0.9051& 0.9971&  0.8267&  0.931\\ \hline 
        PhC-C2DL-PSC& 0.9393& 0.8468&  0.8906& 0.9960&  0.8028&  0.756\\ \hline 
    \end{tabular}
    \label{table4}
\end{adjustwidth}
\end{table}

\section*{Discussion}
The addition of the UNet model to DeepChem, in addition to the improvement in the \textit{ImageLoader}, \textit{ImageDataset} and the \textit{ImageTransformer}, come together to form the Cell segmentation pipeline, which is easy and intuitive to use. Our benchmarking experiments show that the model performs fairly well when compared to the best results on several datasets but still leaves room for improvement. Importantly, the model handled the diverse microscopy techniques and cell samples within the datasets effectively.

\begin{figure}[h]
    \centering
    \includegraphics[width=1\linewidth]{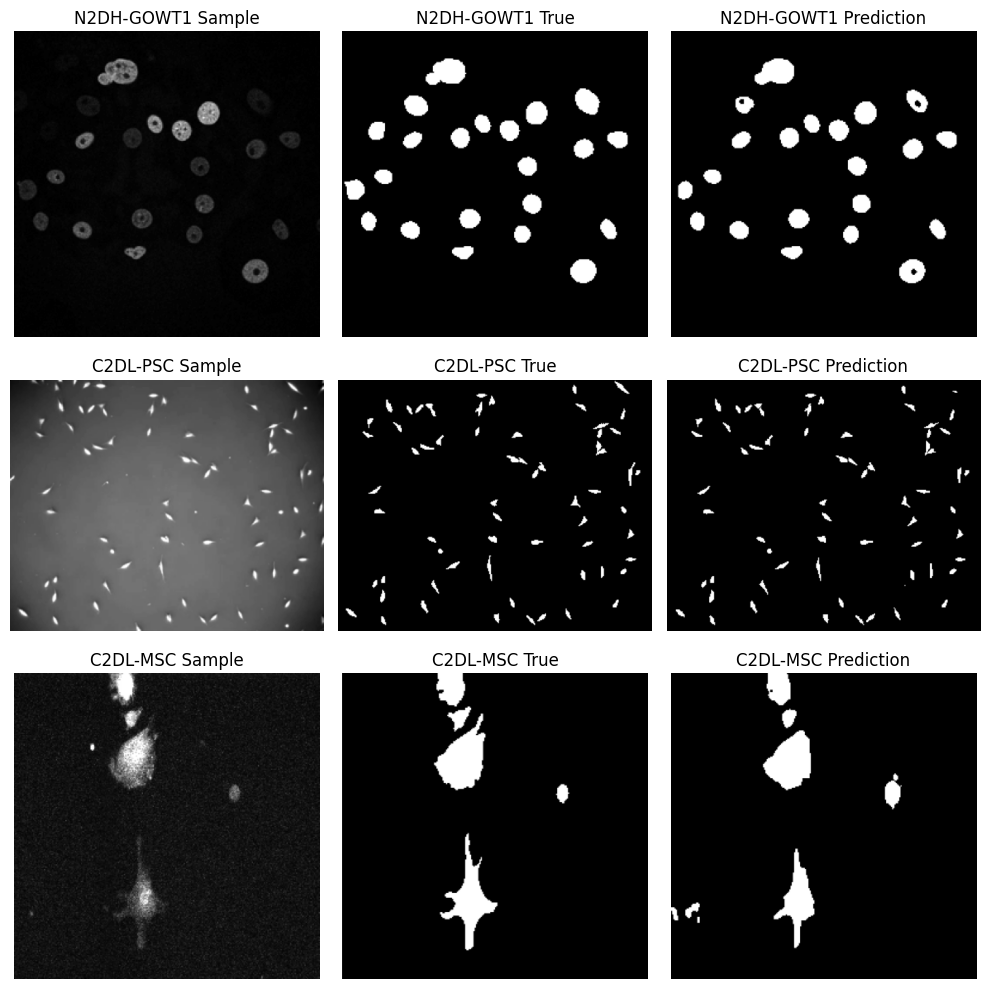}
    \caption{The above image compares the UNet model's predictions with the true segmentation mask. We've compared 1 random sample from the Fluo-N2DH-GOWT1, Fluo-C2DL-MSC and PhC-C2DL-PSC datasets each.}
    \label{figure4}
\end{figure}

\subsection*{Performance and Accuracy}

The UNet model demonstrated robust performance across multiple open-source microscopy datasets. The high F1 Score and Jaccard Index (IoU) achieved in segmentation tasks affirm the model’s capability to handle diverse microscopy techniques and biological subjects. This versatility is critical, as it ensures the model’s applicability across different research domains and imaging modalities, from fluorescence microscopy to DIC microscopy.

\subsection*{Integration with DeepChem and Practical Implications}

Incorporating the UNet model into the DeepChem ecosystem expands the utility of this powerful open-source library. The tutorials and pre-configured pipelines provided within DeepChem facilitate ease of use, making advanced cell segmentation techniques accessible even to researchers with limited expertise in deep learning or image processing. The convenience and ease of use provided by the segmentation pipeline makes it relatively easy to be able to perform segmentation tasks when compared to other frameworks like PyTorch\cite{paszke2019pytorch} or TensorFlow\cite{tensorflow2015-whitepaper}, both of which require extensive knowledge of deep learning to be used effectively. 

One of the primary advantages of our approach is the significant reduction in time and effort required for cell segmentation. Traditional manual segmentation is not only labor-intensive but also subject to human error and variability. By helping to automate this process, our pipeline not only accelerates the workflow but also enhances reproducibility and consistency in segmentation outcomes. This improvement is particularly beneficial for large-scale studies where manual segmentation would be impractical.

In drug discovery, precise cell segmentation can enhance the accuracy of cell counting and morphology analysis, leading to a better understanding of drug effects. In clinical settings, automated segmentation of medical images such as MRIs or X-rays can aid in diagnostics and treatment planning, potentially improving patient outcomes\cite{starkuviene2007potential, wang2021review}.

\subsection*{Limitations and Future Work}

There are several areas for future improvement. Our current implementation and evaluation are limited to a few datasets; expanding this to include a wider variety of datasets could further validate the model’s robustness. Training models on large datasets could help generalize predictions.

Future work could also explore the integration of additional machine learning models within DeepChem to handle other types of biological data, further enhancing its versatility and applicability. Moreover, continuous updates and improvements to the pre-processing and training pipelines could yield even better segmentation accuracy.

\section*{Conclusion}
In conclusion, our work demonstrates a significant step forward in the field of automated cell segmentation. By leveraging the power of the UNet model and integrating it into the user-friendly DeepChem framework, we provide a valuable tool for the scientific community. This advancement not only streamlines cell segmentation tasks but also opens up new possibilities for research and application in various biological and medical fields. The success of this approach highlights the potential of combining deep learning techniques with open-source scientific tools to drive innovation and efficiency in research workflows.


%
%
%

\bibliography{paper_bib}

\end{document}